\documentclass[orivec,runningheads]{llncs}
\usepackage[T1]{fontenc}
\usepackage{cite}
\usepackage{soul}
\usepackage{hyperref}
\usepackage{graphicx}
\usepackage{mdwmath}
\usepackage{amsmath}
\usepackage{amssymb}
\usepackage{booktabs}
\usepackage{tabularx}
\usepackage{array}
\usepackage{verbatim}
\usepackage{xcolor}
\usepackage{multirow}
\usepackage[table]{xcolor}
\usepackage{adjustbox}

\newcolumntype{C}[1]{>{\centering\arraybackslash}p{#1}}

\usepackage{graphicx}
\begin{document}
\title{MammoMix: Leveraging Mixture of Experts for Robust Mammogram Breast Detection}
%
\titlerunning{MammoMix}
%
\author{Dinh Tan Nguyen\inst{1}\orcidID{0000-0002-9749-6924} \and Hoang Quan Dang\inst{2}\orcidID{0009-0000-4218-1389}, 
Chen Zhang\inst{3}\orcidID{0009-0007-0674-8262} \and
Sai Ho Ling\inst{3}\orcidID{0000-0003-0849-5098}}
\authorrunning{Nguyen et al.}
%
\institute{School of Biomedical Engineering, University of Technology Sydney; \\ 
\and
School of Computer Science, University of Technology Sydney; \\
\email{hoangquan.dang@student.uts.edu.au}
\and
School of Electrical and Data Engineering, University of Technology Sydney;\\
\email{\{dinhtan.nguyen, chen.zhang, steve.ling*\}@uts.edu.au}}
\maketitle              
\begin{abstract}


Breast lesion detection in mammography remains a challenging task due to variations in image quality, lesion appearance, and population demographics across datasets. While current object detectors such as YOLO and DETR achieve strong results on individual datasets, their performance often degrades when trained on or applied across heterogeneous sources. To address this, we propose MammoMix, a novel framework based on Mixture-of-Experts (MoE) paradigm for robust and generalizable lesion detection. In MammoMix, each expert model is trained on a specific domain, allowing it to specialize in distinct characteristics of its source data. A gating mechanism adaptively weighs contributions from each expert based on input image, combining their outputs to enable domain-adaptive inference. To improve reliability, we further incorporate a calibration module, MoCAE, which adjusts confidence scores to reflect true predictive uncertainty. We evaluate MammoMix on 3 public mammography datasets: CSAW, DDSM, and DMID, covering diverse clinical settings. Results show that MammoMix outperforms baseline detectors in both average precision and reliability, particularly on datasets with greater variability. Our findings demonstrate that expert specialization and calibrated ensemble fusion significantly enhance model generalization and robustness. MammoMix offers a promising step toward dependable AI-assisted breast cancer screening across real-world clinical domains. Code is available at: \href{https://github.com/tommyngx/MammoMix}{https://github.com/tommyngx/MammoMix}



\end{abstract}
\section{Introduction}

Breast cancer screening via mammography is a critical application of computer vision, where accurate lesion detection significantly influences early diagnosis and outcomes \cite{gotzsche2013screening}. Deep learning-based object detectors such as Faster R-CNN, YOLO, and DETR have shown strong performance on automated mammogram analysis, particularly when trained and tested on a single dataset \cite{giaquinto2022breast,dembrower2020comparison}. In such controlled settings, these models often match or even exceed expert radiologist performance, highlighting their potential for computer-aided diagnosis. However, challenges emerge when models are applied across diverse clinical environments. Real-world mammography data vary significantly across populations, imaging devices, and lesion types. When multiple datasets are combined for training, performance often degrades, even if models perform well on each dataset individually. This drop in generalization is well-documented: detectors that perform strongly on their development set frequently suffer substantial losses in AUC and sensitivity on unseen domains. A unified model struggles to reconcile conflicting data distributions, resulting in unstable training and reduced effectiveness.

To overcome this, we propose MammoMix, a breast lesion detection framework based on a Mixture-of-Experts (MoE) architecture \cite{masoudnia2014mixture}. Instead of a single detector, MammoMix trains multiple expert models, each specialized to a dataset or domain. This divide-and-conquer approach allows each expert to learn domain-relevant features without interference from other distributions. A gating mechanism learns to weigh and combine expert predictions on a per-image basis, allowing the system to dynamically adapt to varying input domains. This specialization improves robustness and generalization across heterogeneous datasets.

Additionally, MammoMix incorporates a confidence calibration module, MoCAE (Mixture of Calibrated Experts), to ensure that expert outputs reflect their actual reliability. Without calibration, overconfident experts may dominate the ensemble, degrading performance. MoCAE calibrates each expert’s confidence score to align predictions with actual precision, enabling fairer fusion. This ensures that the gating mechanism reflects true expertise rather than raw confidence. Together, MoE and MoCAE allow MammoMix to deliver reliable, generalizable lesion detection across diverse mammography data.

\section{Related Work}

Lesion detection in mammograms has attracted significant research interest, with deep learning models increasingly used to support radiologists. Two-stage detectors like Faster R-CNN have shown strong results in identifying masses and calcifications on datasets such as DDSM \cite{kim2020changes}. One-stage YOLO models, valued for their real-time speed, have also demonstrated competitive performance. More recently, Transformer-based detectors \cite{zhang2025ssat} have demonstrated strong potential in the segmentation domain, and architectures such as DETR \cite{carion2020end} have extended this paradigm to object detection, although their application to mammography remains limited \cite{dembrower2020comparison}. DETR introduces self-attention mechanisms to better capture global dependencies, which are essential for detecting subtle lesions in complex mammogram backgrounds. It treats object detection as a set prediction problem, eliminating the need for anchor boxes or NMS by using bipartite matching to align predictions with ground truth\cite{carion2020end}. On the COCO dataset, DETR-ResNet50 achieves 42.0\% mAP, but its slow convergence and computational demands have limited its adoption in real-time medical applications, where faster inference is crucial for clinical workflows. Variants like RT-DETRv2\cite{lv2024rt} address these issues by optimizing for real-time performance, achieving 53.1\% mAP on COCO at 108 FPS on a T4 GPU, outperforming YOLO models or even DETR in both speed and accuracy, and showing promise in medical imaging for tasks like lesion detection in mammograms due to its hybrid CNN-transformer design. However, while these models often achieve competitive accuracy when trained and tested on a single dataset, their performance tends to be domain-specific. Most studies report results confined to the dataset used for training, with accuracy dropping considerably when applied to different sources without adaptation.

Mammography datasets differ by population, imaging equipment, and annotation standards, creating conflicting data distributions that challenge generalization. Multi-center studies \cite{garrucho2022domain} have shown models trained on one dataset perform poorly on others, even with similar imaging protocols. Combining datasets can even reduce sensitivity, as noted by \cite{garrucho2022domain}. In response, domain adaptation methods such as transfer learning and adversarial alignment have been proposed, though they often struggle with overfitting or feature collapse. More recent domain generalization approaches aim to learn transferable features without relying on target data. However, most still rely on a single model stretched across varied data, limiting robustness and motivating a multi-expert strategy.

Mixture-of-Experts (MoE) models \cite{masoudnia2014mixture} assign specialized networks to different sub-tasks or domains, with a gating mechanism selecting the most relevant expert per input. This architecture has re-emerged in modern deep learning as an effective solution for handling diverse data. In medical imaging, MoEs remain underexplored, though ensemble classifiers are sometimes used in breast cancer detection. However, standard ensembles lack learnable gating and treat all models equally, limiting adaptability across domains. MammoMix is among the first to apply MoE specifically for mammographic lesion detection. Recent work on the Mixture of Calibrated Experts (MoCaE) demonstrates that calibration is crucial for effective ensembling \cite{oksuz2023mocae}. MoCaE improves detector fusion by aligning each model’s confidence with its actual performance, avoiding dominance by overconfident predictors. MammoMix builds on this idea by integrating calibration directly into MoE pipeline. Unlike MoCaE’s post-hoc fusion, our experts are trained for domain specialization, optimized jointly with gating and calibration. This enables MammoMix to better accommodate distributional shifts, offering improved accuracy and reliability for multi-dataset breast lesion detection.

\section{Dataset Formulation}

To evaluate the effectiveness of our MoE approach, we employed 3 publicly available mammography datasets: CSAW, DDSM, and DMID. Each dataset brings unique imaging protocols, patient demographics, and lesion types, reflecting the heterogeneity found in real-world screening programs. Training expert models on these distinct subsets allowed us to explore how domain-specific specialization can enhance overall ensemble performance, reduce dataset-specific biases, and improve generalizability across clinical settings.

\subsection{Dataset Description}

The \textbf{Cohort of Screen-Age Women - Case Control (CSAW-CC)} \cite{dembrower2020comparison} contains high-resolution mammograms with detailed pixel-level annotations. Designed to support AI-based breast cancer screening, it includes both cancer and control cases. For this study, we used 472 image-mask pairs, with original image resolutions of 3328×4096 pixels cropped to 1493×3023 to focus on breast regions and improve segmentation efficiency.

The \textbf{CBIS-DDSM (Curated Breast Imaging Subset of DDSM)} \cite{cbisddsmcit} offers 2,620 mammograms from 1,566 patients, paired with pathology reports. We selected 1,244 patients (2,142 images) and focused on 1,333 image-mask pairs involving mass lesions. Images (originally 2986×5491) were cropped to 2134×4627 to standardize input size and enhance training consistency.

The \textbf{DMID (Digital Mammography Dataset)} \cite{oza2024digital} comprises 510 images curated for diagnostic research, with 269 including segmentation masks. Original images (4751×6000) were cropped to 2602×5010 pixels. The dataset provides high-quality DICOM/TIFF images, reports, and pixel-level annotations, supporting breast mass segmentation and model evaluation.

Our expected outcome is enhanced detection accuracy, particularly for under-represented lesion types, as models learn to adapt to inter-dataset variations without assuming homogeneity. The goal of using these datasets instead of a single merged one is to allow our MoE variants to learn specialized representations, simulating real-world clinical scenarios where X-ray images come from varied sources. This mirrors "federated learning" paradigms in healthcare, where data silos are common due to privacy or institutional constraints, aiming to achieve models that are not only accurate but also adaptable to new clinical environments. Quantitatively, the combined dataset size (2,006 images) enables robust statistical evaluation; however, the imbalance (DDSM dominating, fewer positives in DMID) highlights the need for MoE to prevent overfitting to the largest sub-dataset, motivating our designs to pool knowledge across subsets. 

\subsection{Data Preprocessing and Augmentation}

The preprocessing began with image resizing and standardization. All images were resized to a dimension of $640 \times 640$ pixels while preserving aspect ratios through padding, ensuring compatibility with Transformer-based architectures like YOLOS, which benefit from square inputs for efficient attention mechanisms. Following resizing, pixel values were normalized to the range ([0, 1]). Subsequently, we utilized a composition of both geometric and intensity-based augmentations using the albumentations\footnote{https://albumentations.ai} library to simulate real-world variability in mammograms, such as differences in patient positioning, imaging conditions, and tissue density. The goal is to improve model generalization and reduce overfitting by exposing the model to diverse image variations, addressing the imbalance across sub-datasets (e.g., DDSM's dominance) and the scarcity of positive cancer samples. Our geometric augmentations included:

\begin{itemize}
    \item \textit{Elastic Transformations}: Simulate breast tissue deformation (alpha = 50, sigma = 5), helping the model learn deformation-invariant features, especially beneficial for dense breasts.
    \item \textit{Perspective Transform}: Adds mild distortions (scale 0.05–0.1) to mimic imaging angle variations and dataset differences (e.g., DDSM vs. DMID), improving viewpoint robustness.
    \item \textit{Horizontal Flip}: Mirrors images to enforce left-right symmetry learning, enhancing orientation invariance across breast views.
    \item \textit{Rotate}: Small random rotations ($\pm 10^\circ$) simulate patient misalignment while preserving lesion shape, aiding generalization.
    \item \textit{Random Scale}: Varies image size by $\pm 0.2$ to mimic distance and lesion size variability, supporting scale-invariant detection.
    \item \textit{Affine Transformations}: Apply combined scaling, translation ($\pm 0.1$), rotation ($\pm 10^\circ$), and shear ($\pm 6^\circ$) to simulate geometric distortions.
    \item \textit{Random Brightness and Contrast}: Adjusted within $\pm 0.2$ to simulate X-ray exposure variation and enhance robustness to contrast inconsistencies.
    \item \textit{Gaussian Noise}: Added with standard deviation 0.05 to mimic scanner noise and low-quality scans, helping models generalize under real-world artifacts.
    \item \textit{Gaussian Blur}: Slight blur applied to replicate mild defocus, encouraging the model to rely on robust edge features rather than sharp contours alone.
\end{itemize}

All augmentations were implemented with bounding box awareness, ensuring labels were adjusted accordingly using VOC format ($[x_{min}, y_{min}, x_{max}, y_{max}]$) and clipped to image boundaries. Boxes with area less than 25 pixels or visibility below 10\% post-augmentation were discarded to maintain annotation quality. If augmentations resulted in no valid boxes, we re-sampled the image to ensure at least one valid detection target, addressing the risk of losing positive samples in small datasets like DMID. Augmentations were applied only during training without altering validation or test data to reflect real-world inference.

\section{Mixture of Approaches: MoMo, MoE, and MoCAE}

We designed and evaluated 3 approaches using YOLOS as the backbone: MoMo, Simple MoE, and MoCaE. Each variant explores a distinct strategy for leveraging the sub-datasets' heterogeneity, aiming to enhance detection accuracy and robustness by combining the strengths of multiple models or training paradigms. The rationale behind these variants is to exploit dataset-specific patterns while mitigating issues like domain shifts and sub-dataset imbalance. Their designs were motivated by the hypothesis that leveraging dataset diversity through specialized models could outperform a unified model. Our goal is to achieve a detection system that maximizes mAPs across varied clinical scenarios, particularly for small and medium-sized lesions, which are critical for early diagnosis.

\label{sec:Met}
\begin{figure*}[t]
\centering
\includegraphics[width=1.0\textwidth]{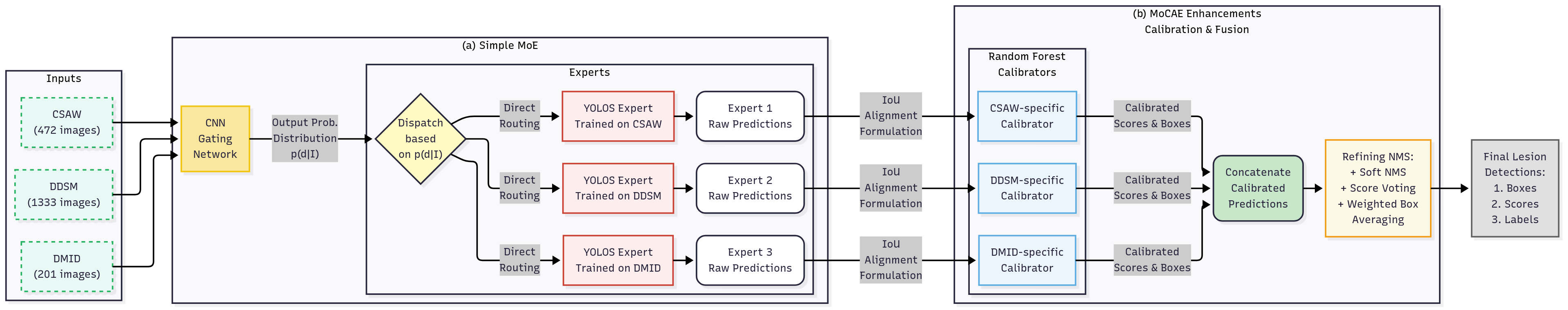}
\caption{Overview of the MammoMix framework: (a) Expert models specialized on individual datasets got selected by a CNN gate (b) Calibration module aligns image embeddings and confidence scores with IoU by employing the predictive reliability from Random Forest; then fusion via Soft NMS and Score Voting.}
\end{figure*}

\subsection{Expert Selection}

We focused on Transformer-based architectures and the core of our detection framework is the YOLOS model\cite{fang2021you}, a Vision Transformer (ViT)\cite{dosovitskiy2020image} object detector that serves as the backbone for all approaches evaluated in this study: MoMo, Simple MoE, and MoCaE. We selected YOLOS over DETR \cite{carion2020end}, despite both being Transformer-based, for several reasons informed by our experiments and literature. First, YOLOS’s purely ViT-based backbone eliminates the need for a convolutional backbone (e.g., ResNet in DETR), reducing architectural complexity of Transformer, which is advantageous for training on moderate-sized datasets like ours (2006 total images). This choice was evidenced by our initial experiments, showing YOLOS outperforming DETR in terms of convergence speed and accuracy on our datasets. YOLOS's novelty in medical imaging, underexplored compared to CNNs or DETR, further motivated its selection, aiming to contribute a new benchmark for Transformer-based detection in mammography.

Unlike traditional CNN-based detectors like YOLO variants that rely on local receptive fields, YOLOS leverages self-attention mechanisms to model long-range dependencies without the inductive biases of convolutions, which is particularly advantageous for medical imaging where contextual relationships, such as the surrounding tissue density influencing lesion detectability, are crucial. The YOLOS architecture applied a simple ViT directly to image patches and treated them as a sequence of tokens, augmented with learnable detection tokens and processed through Transformer layers to predict bounding boxes and class probabilities. Specifically, an input image \(I \in R^{H \times W \times C}\) is divided into non-overlapping patches of size \(P \times P\), flattened into a sequence of \(N \) tokens, where \(N = (H \cdot W) / P^2\). In other words, all images in our datasets were reshaped to \(640 \times 640 \times 3\), resulting in \(N = \lfloor \frac{640}{16} \rfloor^2 = 1600\) patches (patch size \(16 \times 16\)), each embedded into a \(d=768\)-dimensional vector. These patch embeddings were augmented with positional encodings and concatenated with a set of learnable [DET] tokens (fixed at 100), equivalent to "object queries" in DETR, forming the input sequence to the Transformer encoder. The encoder processes this sequence through 12 Transformer layers of multi-head self-attention and feed-forward networks, producing a sequence of contextualized features. The detection heads then predict up to 100 boxes and their class probabilities (in our case, for "cancer" class, plus a "no-object" class) from detection tokens' outputs. 

The output of YOLOS is a set of predictions \(\{(b_i, s_i, l_i)\}_{i=1}^{100}\), where \(b_i \in [0,1]^4\) represents the normalized bounding box coordinates \([x_{\text{center}}, y_{\text{center}}, w, h]\), \(s_i \in [0,1]\) is the confidence score, and \(l_i \in \{0, 1\}\) is the class label (0 for "cancer", 1 for "no-object"). These predictions are matched to ground truth boxes using bipartite matching, optimized via a combination of classification loss (cross-entropy) and box regression loss (L1 and generalized IoU loss). This architecture's simplicity with no anchor boxes or Non-maximum Suppression (NMS) required during single-model inference made it an ideal candidate for adaptation into our MoE frameworks where multiple model outputs must be combined.

\subsection{MoMo: Combined Dataset Training}

The MoMo (Monolithic Model) approach serves as a baseline to assess the efficacy of combining all sub-datasets into a unified source of mammography images without distinguishing their origins. In this method, a single YOLOS model is trained on the aggregated dataset comprising CSAW, DDSM, and DMID. The rationale for MoMo was to test whether a single, general-purpose exposed to a larger, more diverse dataset could generalize across the varied imaging conditions without requiring explicit specialization. By pooling the data, we aimed to maximize training sample size, potentially mitigating the data scarcity issue in smaller datasets like DMID while allowing the model to learn shared features or a robust representation of cancerous lesions across diverse imaging modalities (e.g., masses, calcifications) without the need for specialized training. The goal was to achieve robust performance across all test sets, particularly for DDSM, which dominates the combined dataset. However, we anticipated potential drawbacks: the model might dilute sub-dataset-specific patterns and overfit to DDSM's characteristics, such as its lower-resolution digitized film images, potentially reducing sensitivity to the high-contrast lesions in CSAW or subtle calcifications in DMID. The MoMo approach thus provides a point of comparison to evaluate whether specialized models offer significant advantages over naive data aggregation.

Predictions in MoMo are generated directly by the single YOLOS model, with no additional combination or calibration steps. The output bounding boxes and confidence scores for the "cancer" class are post-processed using standard non-maximum suppression (NMS) to eliminate redundant detections, with an IoU threshold of 0.5 to balance precision and recall. The anticipated outcome is a model that performs adequately across all sub-datasets but may underperform on tasks requiring fine-grained adaptation to dataset-specific nuances, particularly for the smaller DMID dataset, where specialized knowledge is advantageous.

\subsection{Simple MoE: Simple Ensemble of Experts}

The Simple MoE approach trains 3 separate YOLOS models, each as an "expert" on one of the sub-datasets, and combines their predictions using a learned gating mechanism. Unlike traditional MoE implementations that use a learned gating function trained jointly with experts, we trained a separate CNN on the combined dataset to classify an image's dataset origin or to select the most relevant expert for each image. This gating network, denoted as \( G: \mathcal{X} \to \{1, 2, 3\} \), where \(\mathcal{X}\) is the image space and the output indices correspond to CSAW, DDSM, or DMID experts, assigns each test image to the expert most likely to perform well based on its dataset origin. Formally, for an input image \( x \), the gating network outputs a probability distribution over the 3 datasets, \(p(d | I) \in [0,1]^3\), where \(d\) is the dataset and \(I\) is the input image. 

During inference, we select the expert corresponding to the dataset with the highest probability, \(\arg\max_d p(d | I)\), and use its predictions as the final output \(\hat{y}_d = f_d(x)\), where \( f_d \) is the YOLOS model trained on the \( d \)-th sub-dataset. This approach was chosen to simplify the combination process and exploit the specialization of each expert while introducing a dynamic selection mechanism to route images to the most appropriate model, expecting improved performance over MoMo by prioritizing dataset-relevant predictions. However, it risks suboptimal performance if the gating network misclassifies the dataset or if multiple experts produce overlapping or complementary detections, motivating the more sophisticated MoCaE approach.

\subsection{MoCaE: Mixture of Calibrated Experts}

The MoCAE approach, inspired by \cite{oksuz2023mocae}, extends the Simple MoE by incorporating calibration of expert predictions and a sophisticated prediction fusion strategy known as Refining NMS to address the limitations of naive ensemble methods. This approach addressed the critical issue of miscalibration in ensemble methods, where raw confidence scores from different experts may be misaligned due to variations in training data, leading to biased combinations where a confident but inaccurate expert dominates. Calibration aligns confidence scores with the true likelihood of correct detection, measured as the maximum IoU with ground truth boxes. Formally, for a prediction with confidence score \( s \), the calibrator \( C: s \to \hat{s} \) should map \( s \) to an expected IoU value, such that \(\hat{s} \approx E[\text{IoU}(b, g) | s]\), where \( b \) is the predicted box and \( g \) is the ground truth. We aim to maximize the complementary strengths and ensure fair contribution from each expert, expecting improved overall mAP, particularly for challenging cases like small and medium-sized lesions where dataset-specific expertise is critical.


For each YOLOS expert, we constructed a calibration dataset using validation images from all 3 sub-datasets. This was motivated by the need to capture both in-domain and out-of-domain performance, as experts may encounter images from other datasets during testing, reflecting real-world scenarios where image sources are mixed. For an expert \( e \) trained on dataset \(D_e\) (e.g., CSAW), we processed validation images from all datasets to collect:
\begin{enumerate}
    \item Image embeddings extracted using a pre-trained ResNet-18 backbone (without its classification head), producing a 512-dimensional feature vector per image, capturing high-level visual characteristics like texture and contrast.
    \item Confidence scores from the expert's predictions for the "cancer" class.
    \item Intersection over Union (IoU) values: For in-domain images from \(D_e\), we computed predictions \(\{(b_i^e, s_i^e, l_i^e)\}\) and their maximum IoUs with ground truth boxes, \(\text{IoU}_{\max} = \max_{k} \text{IoU}(b_i^e, g_k)\), where \(g_k\) is a ground truth box. For out-of-domain images from other datasets (\(D_k, k \neq e\)), we assign an IoU of 0, as these images lack ground truth annotations relevant to the expert's training domain, reflecting the expert's likely poor performance due to domain mismatch. The confidence score \(s_i^e \in [0, 1] \) and image embeddings \( f \in \mathbb{R}^{512} \) form the input, resulting in a \(513\)-dimensional vector and paired with target \(\text{IoU}_{\max} \in [0, 1]\) for in-domain or 0 for out-of-domain samples.
\end{enumerate}

3 separate calibrators were trained, one per expert, to account for dataset-specific biases and ensure each expert’s predictions are adjusted independently. Here, we employed 3 Random Forest regressors as 3 calibrators to minimize the Mean Squared Error between predicted and actual IoU values: \(\text{MSE} = \frac{1}{N} \sum_{k=1}^N \left( C([f_k, s_k]) - \text{IoU}_{\max,k} \right)^2\), where \( N \) is the number of validation samples, and \( C \) is the calibrator. The Random Forest was chosen over isotonic regression (used in the original MoCaE\cite{oksuz2023mocae}) due to its robustness to high-dimensional inputs and ability to model non-linear relationships between embeddings, scores, and IoU, outperforming simpler methods when incorporating image features. This calibration strategy ensures the calibrator learns to assign low confidence to out-of-domain predictions, preventing overconfident errors. We aimed to mimic clinical settings where image sources are not pre-identified, enhancing the generalizability and reliability of confidence scores in mixed-domain testing scenarios.

During inference, all 3 experts process each test image, producing predictions \(\{(b_i^e, s_i^e, l_i^e)\}_{i=1}^{100}\). Confidence scores are calibrated using the respective Random Forests, yielding \(\tilde{s}_i^e \approx \text{IoU}(b_i^e, g_k)\). These predictions are concatenated into a single set, filtered for the "cancer" class (\(l_i^e = 0\)), and processed through Refining NMS, which acted as a routing mechanism to handle overlapping predictions from multiple experts with 2 stages:
\begin{itemize}
    \item Soft NMS: Unlike traditional NMS, which discards overlapping boxes based on a fixed IoU threshold, Soft NMS reduces the scores of overlapping boxes using a Gaussian decay. For a box \( b_i \) with calibrated score \( \tilde{s}_i \), and another box \( b_j \) with \(\text{IoU}(b_i, b_j) > \theta\), the score is updated as \( \tilde{s}_i \gets \tilde{s}_i \cdot \exp\left(-\frac{\text{IoU}(b_i, b_j)^2}{\sigma_{\text{nms}}}\right) \), where \(\sigma_{\text{nms}}\) controls the influence of nearby boxes. This soft suppression, inspired by the MoCAE framework in \cite{oksuz2023mocae}, was chosen to reduce false negatives and avoid hard suppression of valid detections, particularly for dense or clustered lesions common in mammography.
    \item Score Voting: After Soft NMS, surviving boxes are refined by weighted averaging based on their calibrated scores and pairwise IoUs, where the weight for box \(b_i\) is computed as \(w_{ij} = \tilde{s}_i \cdot \exp\left(-\frac{(1 - \text{IoU}(b_i, b_j))^2}{\sigma_{\text{nms}}}\right)\), excluding self-influence (\(i \neq j\)). The refined box is \( b_i' \) = \(\sum_{j \neq i} w_{ij} \cdot b_j / \sum_{j \neq i} w_{ij}\). This fusion improved box localization by aggregating information from all experts to refine their coordinates, leveraging their complementary detections strengths
\end{itemize}

\section{Experimental Setup}\label{ES} 


YOLOS with base version serves as the unified backbone across all MoE approaches, ensuring consistency in feature extraction while allowing each variant to explore different training and combination strategies. Optimization used AdamW with a learning rate of \(5 \times 10^{-5}\), weight decay of \(10^{-4}\), and a cosine scheduler with warmup ratio 0.05 and one restart cycle. The batch size was set to 8, with gradient accumulation over 2 steps to simulate larger effective batches on a 40GB memory NVIDIA A100 GPU. We trained for 200 epochs, monitoring validation mAP to enable early stopping if no improvement occurred for 20 epochs, preventing overfitting on small datasets like DMID. This extended epoch count was selected to allow sufficient adaptation of the transformer layers, which often require longer training than CNNs due to their data-hungry nature.  For the MoCAE variant, post-training calibration was integrated using a Random Forest regressor with 300 estimators to predict maximum IoU values, producing calibrated scores that better reflect true detection quality to enable more effective combination with a \(\sigma_{\text{nms}} = 0.08\) in the ensemble phase. 

The splitting strategy was chosen to allocate approximately 64\% of data for training, 16\% for validation (to tune hyperparameters and calibrate models), and 20\% for testing (to evaluate generalization), as shown in Table~\ref{tab1}. The validation set is particularly important for MoCAE, where it is used to fit 3 calibrators associated with 3 datasets used, ensuring that confidence scores are aligned with actual detection quality. 

\begin{table}
\centering
\caption{Data distribution across training, validation, and test sets}\label{tab1}
\begin{tabular}{C{2cm} C{2cm} C{2cm} C{2cm} C{2cm}}
\toprule
Subset & Train & Validation & Test & Total\\
\midrule
CSAW & 301 & 76 & 95 & 472\\
DDSM & 852 & 214 & 267 & 1333\\
DMID & 128 & 32 & 41 & 201\\
\bottomrule
\end{tabular}
\end{table}

In evaluating the performance of our methods, several key metrics are used to measure accuracy and overlap between predicted and true regions. Among these metrics, the Mean Average Precision (mAP) \cite{hamed2020deep} is a commonly used measure, especially in object detection and segmentation tasks. The mAP values are provided for different IoU thresholds to capture the performance across various levels of overlap criteria. The IoU=0.50 (average) measures precision when the IoU between predicted and ground truth regions is at least 0.50. The IoU=0.50:0.95 (average) calculates average precision across multiple IoU thresholds, offering a comprehensive performance overview. The IoU=0.50:0.95 (large) focuses on larger objects, while IoU=0.50:0.95 (medium) targets medium-sized objects, and IoU=0.50:0.95 (small) evaluates performance for smaller objects, providing insights into detection and segmentation accuracy for different sizes.

\section{Results and Discussion}

Our preliminary experiments confirmed that YOLOS-base outperformed DETR on mAP@50 across datasets, likely due to its simpler design and fewer modifications needed for fine-tuning on Table~\ref{tab2}. On the CSAW dataset, individually trained YOLOS experts achieved the highest mAP@50-95 (0.3292), outperforming DETR (0.2763), RT-DETRv2 (0.2260), and even the ensemble-based approaches. While Simple MoE reached comparable performance (0.3287), the calibration-enhanced MoCAE showed slightly lower overall mAP (0.3044), yet it achieved the best performance for small lesions (0.3119). This suggests that calibration improves model reliability in detecting subtle and low-contrast abnormalities, which are typically more challenging in clinical settings.

\begin{table}[htbp]
\caption{Comparison of mAPs for different models across 3 datasets}\label{tab2}
\begin{tabular}{|c|c|cc|cccc|}
\hline
\rowcolor{blue!15}
\textbf{Subset} & \textbf{mAP} & \multicolumn{2}{c|}{\textbf{DETR baseline}} & \multicolumn{4}{c|}{\textbf{YOLOS}} \\
\rowcolor{blue!15}
 &  & DETR & RT-DETRv2 & Individually & All (MoMo) & Simple MOE & MoCAE \\
\hline
\multirow{6}{*}{\textcolor{black}{\textbf{CSAW}}} 
& 50-95   & 0.2763 & 0.2260 & \textbf{0.3292} & 0.2458 & 0.3287 & 0.3044 \\
& 50      & 0.6491 & 0.4687 & \textbf{0.7185} & 0.6321 & \textbf{0.7185} & 0.6854 \\
& 75      & \textbf{0.1616} & 0.1250 & 0.1524 & 0.1149 & 0.1481 & 0.1161 \\
& small   & 0.2673 & 0.1181 & 0.2106 & 0.1641 & 0.2106 & \textbf{0.3119} \\
& medium  & 0.2683 & 0.2000 & \textbf{0.3210} & 0.2568 & \textbf{0.3210} & 0.2978 \\
& large   & 0.3942 & 0.4325 & \textbf{0.4932} & 0.2966 & 0.4881 & 0.4447 \\
\hline
\multirow{6}{*}{\textcolor{black}{\textbf{DDSM}}} 
& 50-95   & \textbf{0.1995} & 0.0720 & 0.1973 & 0.1710 & 0.1973 & 0.1357 \\
& 50      & 0.5027 & 0.1813 & \textbf{0.5103} & 0.4699 & 0.5047 & 0.3508 \\
& 75      & \textbf{0.1042} & 0.0410 & 0.0989 & 0.0810 & 0.0972 & 0.0562 \\
& small   & \textbf{0.2901} & 0.1765 & 0.2066 & 0.2024 & 0.2316 & 0.1975 \\
& medium  & \textbf{0.2036} & 0.0673 & 0.1986 & 0.1658 & 0.1986 & 0.1318 \\
& large   & 0.1892 & 0.0965 & \textbf{0.1998} & 0.1975 & 0.1927 & 0.1576 \\
\hline
\multirow{6}{*}{\textcolor{black}{\textbf{DMID}}} 
& 50-95   & 0.1965 & 0.2030 & \textbf{0.3364} & 0.2660 & 0.3348 & 0.3204 \\
& 50      & 0.4180 & 0.3456 & \textbf{0.6249} & 0.5630 & \textbf{0.6249} & 0.6016 \\
& 75      & 0.1360 & 0.2210 & 0.2409 & 0.2299 & 0.2409 & \textbf{0.2464} \\
& small   & 0.1010 & 0.3535 & 0.0000 & \textbf{0.4040} & 0.0000 & 0.0505 \\
& medium  & 0.1519 & 0.1225 & 0.2437 & 0.1488 & 0.2437 & \textbf{0.2504} \\
& large   & 0.3297 & 0.3086 & \textbf{0.4805} & 0.4169 & 0.4763 & \textbf{0.4910} \\
\hline
\end{tabular}
\end{table}

For the DDSM dataset, the DETR baseline outperformed all other approaches in terms of mAP@50-95 (0.1995), with YOLOS individually (0.1973) and Simple MoE (0.1973) closely following. This indicates that for more homogeneous datasets like DDSM, a strong unified model such as DETR can still generalize effectively without requiring domain-specific specialization. Nevertheless, MoE variants remained competitive across lesion sizes, particularly for large lesions, maintaining stable performance without significant degradation.

In contrast, on the DMID dataset, which is smaller and more diverse, the MoE-based models demonstrated clear advantages. The YOLOS individually trained experts reached the highest mAP@50-95 (0.3364), while both Simple MoE (0.3348) and MoCAE (0.3204) followed closely. Notably, MoCAE achieved the highest mAP for large lesions (0.4910) and partially recovered detection capability for small lesions (0.0505), where YOLOS individually failed (0.0000). This highlights MoCAE’s ability to provide more balanced and reliable predictions across lesion scales.

\begin{figure*}[t]
\centering
\includegraphics[width=0.99\textwidth]{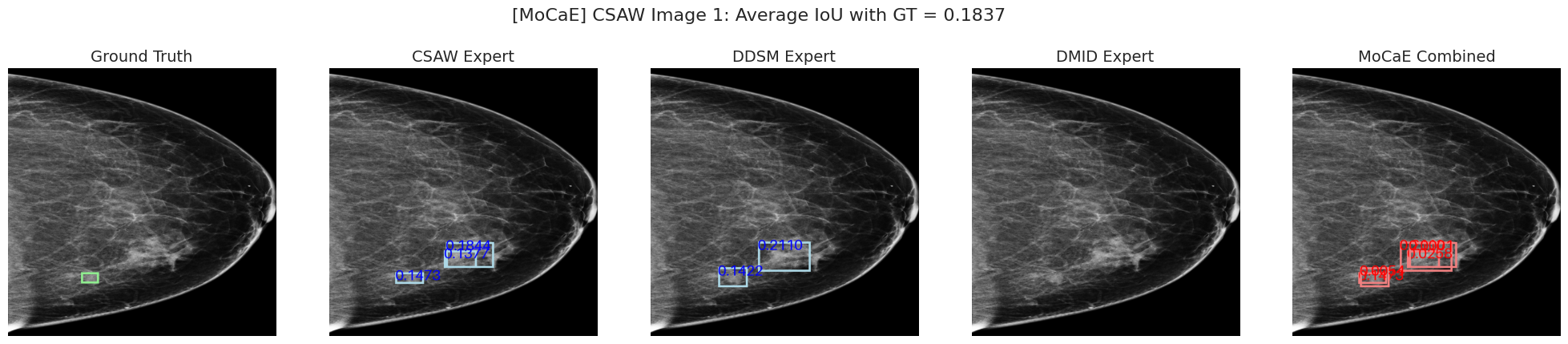}
\includegraphics[width=0.99\textwidth]{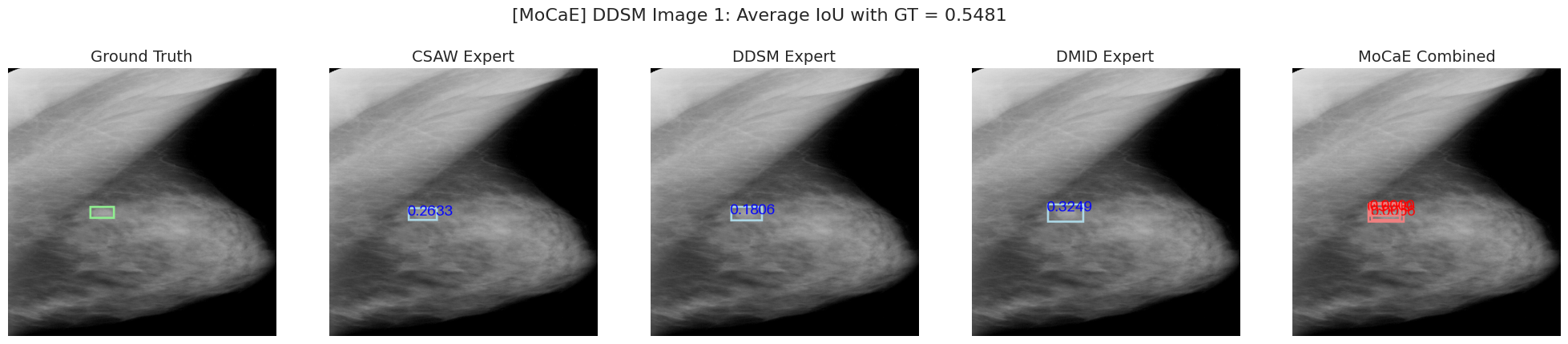}
\includegraphics[width=0.99\textwidth]{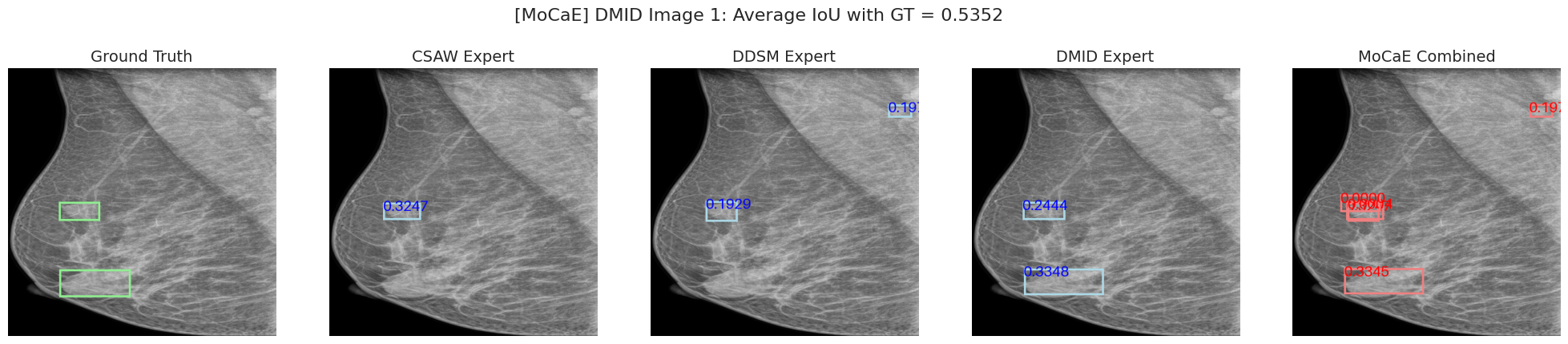}
\caption{Inference Breakdown Across Multiple Expert Models accross dataset}\label{example2}
\end{figure*}

Figure~\ref{example2} illustrates how the MoE-based combination (MoCaE Combined) consistently outperforms individual experts in terms of localization accuracy across diverse datasets. In each row, we observe that while individual experts (trained on CSAW, DDSM, or DMID) produce fragmented or misaligned detections—often with low IoU scores—the combined MoCaE output yields more complete and precise bounding boxes that better match the ground truth. For instance, in the CSAW image (top row), none of the individual experts align well with the lesion, whereas the MoCaE fusion successfully captures the region with higher spatial coverage. Similarly, in the DDSM and DMID examples (second and third rows), the MoCaE output consolidates partial detections from different experts into a unified and more accurate prediction. These qualitative results support the advantage of expert specialization and calibrated fusion in handling domain-specific variance and improving overall lesion detection.

While our framework improved robustness in breast lesion detection across diverse mammography datasets, several limitations warrant consideration. First, computational costs remain a concern. Training multiple experts (3 YOLOS-base models, each with 127.73M parameters, over 200 epochs) obviously requires substantial resources comparing to a single one, with total training time exceeding 4 GPU-hours on a single NVIDIA A100, which will be potentially longer on larger dataset. Inference runtime, while moderately efficient at around 8 seconds per image for the ensemble, is 2–3 times slower than a single YOLOS model due to the calibration and fusion steps (e.g., Soft NMS and Score Voting), potentially hindering deployment in resource-limited clinical settings where real-time processing is essential. Moreover, clinical interpretability poses another challenge. Although MoCAE enhances confidence calibration, the "black-box" nature of YOLOS transformers limits radiologists' ability to understand decision-making processes, such as how attention mechanisms prioritize subtle lesions. Without integrated explainable AI (XAI) tools like Grad-CAM, the system may face resistance in clinical adoption, where transparency is critical for trust and error tracing. Dataset biases also further complicate generalization. Our subsets (CSAW, DDSM, DMID) exhibit imbalances (e.g., DDSM's 1,333 samples vs. DMID's 201), potentially amplifying demographic or imaging protocol biases common in mammography data, such as underrepresentation of dense breasts or diverse ethnic groups. This could lead to reduced performance on underrepresented cases, as evidenced by MoCaE's lower mAP on DDSM large objects, reflecting domain shifts not fully addressed by our calibration.

Overall, these results support the effectiveness of the Mixture-of-Experts approach in addressing domain heterogeneity, particularly in datasets with diverse characteristics like CSAW and DMID. While unified models may suffice for cleaner, single-domain datasets, MoE models—especially when enhanced with confidence calibration via MoCAE—offer improved robustness, generalization, and reliability across varied clinical scenarios. This makes them promising candidates for real-world deployment in breast cancer screening workflows.

\begin{credits}

\subsubsection{\ackname} We acknowledge the support from the Research team of the University of Technology Sydney for providing storage and facilities to conduct this study.

\section{Conclusion}

In this work, we introduced MammoMix, a Mixture-of-Experts framework designed to improve breast lesion detection across heterogeneous mammography datasets. By training specialized experts on individual data domains and integrating their outputs through a learned gating mechanism, MammoMix effectively addresses domain-specific variability. We further incorporated a calibration module, MoCAE, to ensure reliable and well-calibrated predictions. Experimental results across three diverse datasets, CSAW, DDSM, and DMID, demonstrated that MammoMix consistently improves detection accuracy and robustness, particularly in challenging or mixed-domain settings. These findings highlight the potential of MoE-based architectures in advancing generalizable, trustworthy AI tools for clinical breast cancer screening. Our future iterations on this matter could incorporate lightweight architectures or pruning to reduce costs, while integrating XAI for better interpretability.

\subsubsection{\discintname}

\footnote{None. Submission to AJCAI2025 with original research.}

\end{credits}
%
%
%
%

\bibliographystyle{splncs04}
\bibliography{refs}





\end{document}